\documentclass[11pt,a4paper]{article}
\usepackage{times,latexsym}
\usepackage{url}
\usepackage[T1]{fontenc}

\usepackage[acceptedWithA]{tacl2021v1}

\usepackage{xspace,mfirstuc,tabulary}

\newif\iftaclinstructions
\taclinstructionsfalse %
\iftaclinstructions
\renewcommand{\confidential}{}
\renewcommand{\anonsubtext}{(No author info supplied here, for consistency with
TACL-submission anonymization requirements)}
\newcommand{\instr}
\fi

\iftaclpubformat %

\else

\fi

\usepackage{svg}
\usepackage{amsmath}
\usepackage{mdframed} %
\usepackage{booktabs} %
\usepackage{multirow} %
\usepackage{dsfont} %
\usepackage{tikz}
\usepackage{mathtools}
\usepackage{nicematrix}

\usepackage{array,multirow,graphicx}
\usepackage{float}

\usepackage{caption}
\usepackage{subcaption}

\newcommand{\cvar}[1]{\textsc{#1}}

\newcommand{\cex}[1]{\texttt{#1}}

\definecolor{superlightred}{HTML}{F5F5F5}
\newmdtheoremenv[backgroundcolor=superlightred,
frametitlerulewidth=0pt,
innertopmargin=3pt,
innerrightmargin=3pt, 
innerleftmargin=3pt, 
innerbottommargin=3pt
]{hypothesis}{Hypothesis}

\title{Measuring Causal Effects of Data Statistics on Language Model's `Factual' Predictions}

\author{Yanai Elazar\textsuperscript{1,2} \,
 Nora Kassner\textsuperscript{3} \,
 Shauli Ravfogel\textsuperscript{1,2} \, \\
 {\bf Amir Feder\textsuperscript{4} \,
 \bf Abhilasha Ravichander\textsuperscript{5} \,
 \bf Marius Mosbach\textsuperscript{6} \,
 }\\
 {
 \bf Yonatan Belinkov\textsuperscript{4}\,
 \bf Hinrich Sch\"utze\textsuperscript{3}\, 
 Yoav Goldberg\textsuperscript{1,2}}\\
\textsuperscript{1}Bar-Ilan University,
\textsuperscript{2}Allen Institute for Artificial Intelligence \\
\textsuperscript{3}CIS, LMU Munich,
\textsuperscript{4}Technion,
\textsuperscript{5}LTI, Carnegie Mellon University,
\textsuperscript{6}Saarland University\\
{\tt  yanaiela@gmail.com}\\
  }

\date{}

\begin{document}
\maketitle
\begin{abstract}

Large amounts of training data are one of the major reasons for the high performance of state-of-the-art NLP models.
But what exactly in the training data causes a model to make a certain prediction? 
We seek to answer this question by providing a language for describing how training data influences predictions, through a causal framework.
Importantly, our framework bypasses the need to \emph{retrain expensive models} and allows us to estimate causal effects based on observational data alone.
Addressing the problem of extracting factual knowledge from pretrained language models (PLMs), we focus on simple data statistics such as co-occurrence counts and show that these statistics do influence the predictions of PLMs, suggesting that such models rely on shallow heuristics.
Our causal framework and our results demonstrate the importance of studying datasets and the benefits of causality for understanding NLP models.
\end{abstract}

\section{Introduction}

\begin{figure}[t!]
\centering
\includegraphics[width=1.\columnwidth]{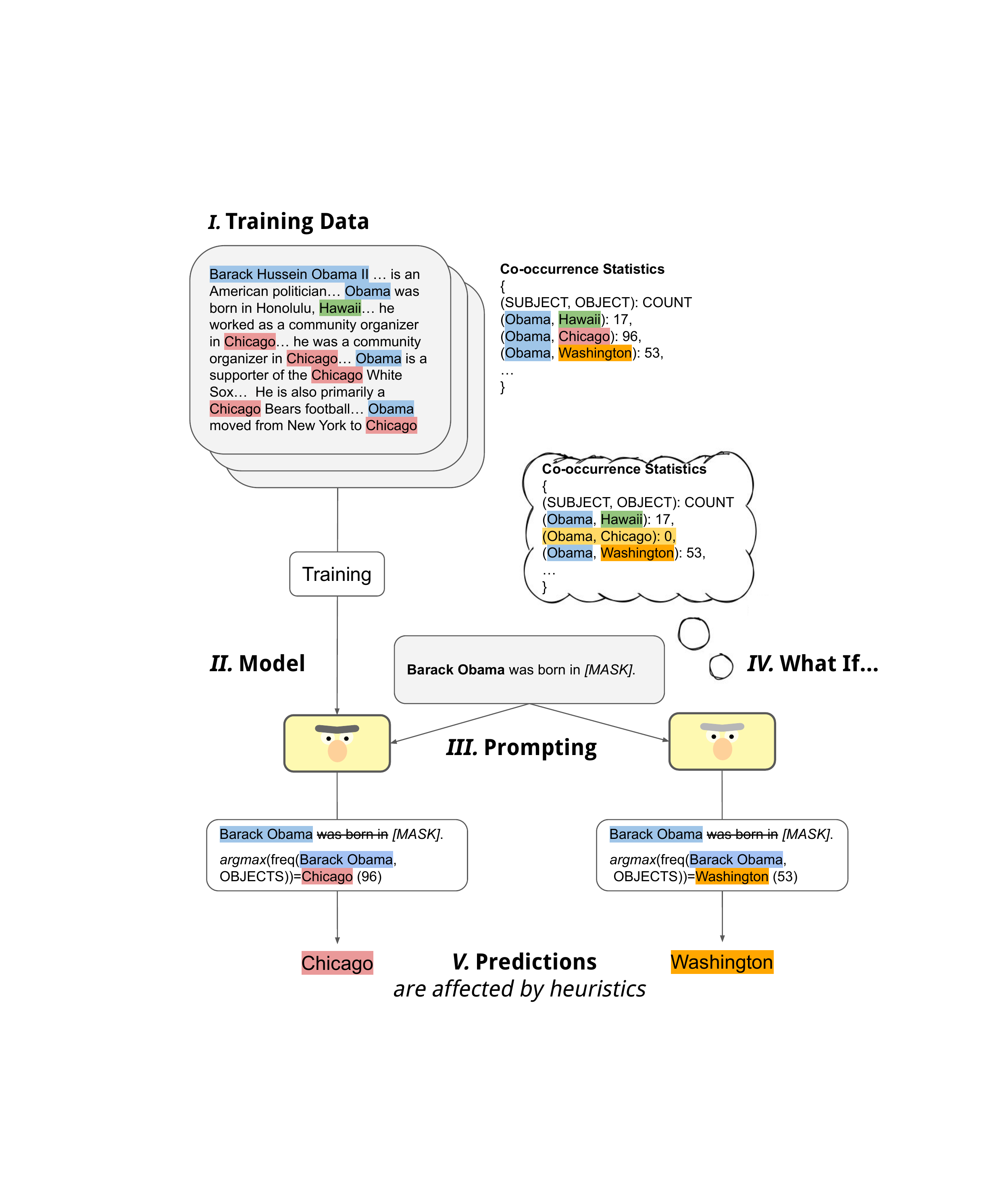}
\caption{
In this figure, we showcase one of our hypotheses of how the model is influenced by the co-occurrences between subjects and objects in the training data, while ignoring the meaning of the pattern itself. 
While in the original training data the most co-occurring object with \textit{Barack Obama} is \textit{Chicago}, in the simulated world (which we estimate through observational data) it is \textit{Washington}, which is the updated model's prediction.}
\label{fig:intro_fig}
\vspace{-0.2in}
\end{figure}

To what extent are predictions of neural language-models 
influenced by simple statistics in the training data, rather than by deeper understanding of the text? For example, when asked to complete ``Barack Obama was born in [MASK]'', is the model just choosing a location which frequently co-occurred with Barack Obama in training, while ignoring the word ``born''? We often suspect that models indeed make such shortcuts based on the training data \cite{gururangan2018annotation,poliak2018hypothesis,tsuchiya2018performance,naik-etal-2018-stress,kaushik2018much,elazar2021back}. Understanding such mechanisms is essential for better model interpretation and analysis.  
However, this behavior, often backed up by corpus counts \cite{razeghi2022impact}, only reflects a \emph{correlation} between corpus statistics and model predictions, and correlation does not imply causation. 

In this work, we attempt to formalize these intuitions and show how to make more rigorous, \emph{causal} claims. The classic way to show causation is using intervention: in our case, re-training the model on a training dataset with altered co-occurrence statistics, and then to measure the effect on its predictions. With pretrained language models (PLMs) and large training corpora, this is highly impractical. Instead, we show how we can use causal inference to make causal claims based on observational data alone.
While the tools we use for making causal claims from observational data---causal graphs, do-calculus, and backdoor criterion---are well established \cite{pearl}, their application to the problem of quantifying the influence of training data on model behavior is both novel and nontrivial, and most of this paper is dedicated to the details of this application.
We use the causal framework to analyze three different co-occurrence based heuristics, and our results are evidence that PLMs indeed causally use them while making predictions.
An illustrative example of one such hypothesis is presented in Figure \ref{fig:intro_fig}.

Concretely, our empirical focus is the origin of seemingly \emph{factual knowledge} expressed by models.
We query LMs with prompting for extracting factual knowledge \cite{lama}.
Given a \cex{<subject, relation, object>} triplet from some knowledge base (KB), we construct a \textit{cloze-pattern} (e.g. ``Barack Obama was born in [MASK]'') from the subject and the relation. Then we feed it to a model and record its prediction.
In contrast to more abstract properties such as syntax that
involve generalization over large, diverse sets of
sentences, factual knowledge extraction is more grounded to
specific examples in the training data. The focus on factual knowledge allows us to investigate concrete hypotheses, as well as easily trace relevant information from the training data.

We hypothesize that models acquire shallow heuristics rather than abstract factual knowledge, and while such heuristics may result in above-random performance, reliance of the model on them indicates a lack of generalization capabilities.
We present three hypotheses to explain model behavior in the setup of factual knowledge extraction, related to the training data:
\begin{enumerate}
    \item \textit{Exact-Match}: models memorize utterances from the training data and predict the object that appeared in the original utterance. 
    \item \textit{Pattern-Object Co-occurrence}: models
    predict the object that appears most often with some textual \textit{pattern} that expresses some relation (regardless of the subject).
    \item \textit{Subject-Object Co-occurrence}: models
    predict the object that co-occurs most often with some subject (regardless of the pattern).
\end{enumerate}

For establishing a causal effect, we begin by constructing a causal graph that encodes our assumptions of the different factors and their interconnections (e.g. the training data affects the model, since a change on the former would change the latter as well) (\S \ref{sec:graph}).
Using the graph, together with causal inference techniques, we showcase how to estimate causal effects for each hypothesis (\S \ref{sec:estimation}).
We describe the data collection required for estimating each causal question, estimate the causal effect of each hypothesis, and the models we consider (\S \ref{sec:setup}), and find significant causal effects of the heuristic mechanisms implied by the hypotheses, for the 32 PLMs we consider (\S \ref{sec:restuls}). One of these models include a new RoBERTa-base model \cite{roberta} we train, and save the 84 checkpoints after each epoch, to study the training dynamics with regard to the heuristics, which we release to the community.

Beyond these concrete findings, our contribution is methodological: we believe this work to be just the first step, and that causal frameworks can, and should, be used for analyzing many other causal relations between a model's training data and its predictions.

\section{Background}

\subsection{Causal Analysis Methods in NLP}

\paragraph{Interventions in representation space.} Recent applications of causal approaches for the interpretation of NLP systems aim to \emph{simulate} a controlled experiment by intervening on model's representations. This is typically done by a modification of the encoding of a given human-interpretable concept of a pretrained model, and relating the intervention to the change in the model's behavior~\cite{giulianelli-etal-2018-hood,lakretz2019emergence,bau2018identifying,amnesic_probing,CausaLM,relative-clauses,antverg2022on}. 
In all cases, one aims to create a counterfactual model, which is different than the original one only in the encoding of the concept.  %

\paragraph{Causal-based approaches.} It is not always possible to simulate a counterfactual model by representation-based intervention. For instance, one cannot easily ``erase'' the encoding of a specific subject-object pair from a given pretrained model. Additionally, naively applying concept-based intervention does not generally consider all relevant confounds, and can thus mis-attribute the causal effect of the concept. 
As such, several works have used causal graphs as an interpretation tool. \citet{vig2020causal} have used causal mediation analysis to attribute gender bias to individual model components (neurons and attention heads), and in order to identify the source of that bias. \citet{finlayson-etal-2021-causal} followed a similar framework to investigate mechanisms of subject--verb agreement in language models. 
\citet{slobodkin2021mediators} have identified context length as a mediator in probing-based analysis of the localization of linguistic concepts within the network. Beyond analysis, \citet{wu2021causal} have used an intervention-based method for model distillation. To the best of our knowledge, our work is the first attempt to assess the causal influence of specific properties of the training data on the model's predictions. 

Other work has categorized the causal direction of data collection in different tasks and datasets, and proposed a method based on minimum description length to automatically detect the direction from data \cite{jin2021causal}. Yet other work has sought to characterize spurious features using techniques from causal analysis, by identifying these features as ones that are counterfactually invariant~\cite{veitch2021counterfactual} or for examining existing definitions for spurious features~\cite{eisenstein-2022-informativeness}.
Finally, \citet{wei2021frequency} investigated how word frequency and co-occurrences causally affect subject--verb agreement. In contrast to our work, they performed experiments that involved re-training several models on different data splits, where they controlled the frequency of the investigated terms.
We refer to \citet{feder2021causal} for an overview of causal-based approaches and challenges in NLP.

\subsection{Data as Explanations}

Not much work has been devoted to explain models based on their training data. Perhaps the most relevant, and popular approach to date is \textit{Influence functions} \cite{hampel1974influence,koh2017understanding}. This method approximates the causal effect of a single example from training on a test example, however, they do not provide abstract explanations of model behavior, as we seek in this work. Moreover, interpreting how specific examples influenced a model is challenging.

Recently, \citet{razeghi2022impact} showed that the frequency of numerals in GPT-like models is strongly correlated with their ability to solve simple math problems.
While they focus on co-occurrences, their findings are correlational in nature, rather than the desired, causal explanations. Similarly, other work has constructed datasets to evaluate the extent of memorization by language models~\cite{mccoy2021much,emami-etal-2020-analysis}. 
Closest to our work is that of~\cite{akyurek2022tracing}, which examines techniques to trace language model's factual assertions back to the training data and find that existing influence methods fail to fact trace reliably. In contrast, in our work, we formulate hypotheses for the predictions of language models on prompts requiring factual knowledge, and establish their effect on models' predictions using a causal approach.

\subsection{Language Models as Knowledge Bases}

\citet{lama} first studied the question of 'Language Models as Knowledge Bases?' and showed that knowledge can be directly extracted from LMs without providing any external, explicit knowledge source. %
Subsequent work studied limitations of the LMs-as-KBs paradigm: \citet{poerner-etal-2020-e} point out that performance is due to easy to guess names; \citet{dufter-etal-2021-static} hypothesize that performance is due to similarity assessments on a type-restricted vocabulary much like a nearest neighbor search for static embeddings. \citet{cao-etal-2021-knowledgeable} analyse prompt bias and reliance on typing. Finally, a collection of work studies inconsistency of knowledge captured inside LMs with respect to paraphrased relations \cite{pararel}, negation \cite{ettinger-2020-bert, kassner-schutze-2020-negated}, multilinguality \cite{jiang-etal-2020-x, kassner2020pretrained}, singular and plural hypernymy probes \cite{ravichander-etal-2020-systematicity} and common-sense constraints \cite{kassner-etal-2021-beliefbank}. More general, \citet{DBLP:journals/corr/abs-2110-04888} outlined characteristic differences between LMs and KBs qualitatively.
Our work adds to this discussion by causally connecting model predictions to properties in the training data and therefore explaining LM's behavior.

\section{Causal Graph and Hypotheses for the LMs-as-KBs Setup}

\label{sec:graph}

\begin{figure*}[t]
\centering
\includegraphics[width=.9\textwidth]{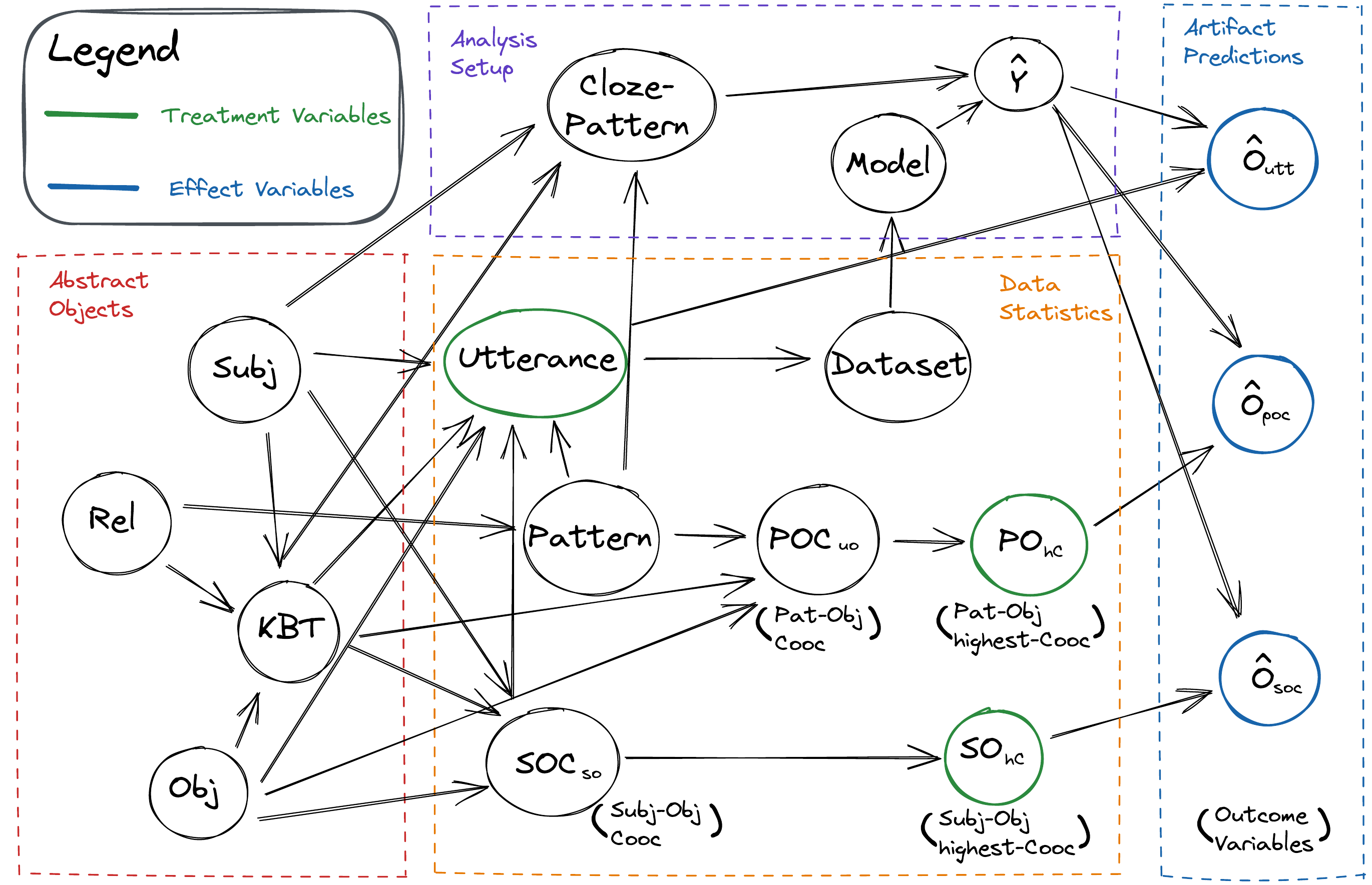}
\caption{Our causal graph; it encapsulates all our assumptions about the causal relationships between the different variables.
The three outcome variables ($\hat{O}_{utt}$,
$\hat{O}_{poc}$, $\hat{O}_{soc}$) match the corresponding
variables whose influence on the prediction
we measure.%
This graph is explained in detail in Section \ref{subsec:graph-explained}.
}
\label{fig:graph}

\end{figure*}

In this section we propose a causal graph (directed acyclic graph) that specifies our presupposed causal relations between different random variables. We construct such graph based on the different properties of the training data, the evaluation setup and the model predictions, and describe it in what follows. 
The nodes in the graph correspond to variables such as an \cvar{utterance}, or a \cvar{prediction}, while the edges define the \textit{causal relationships} between the variables. Any missing edge indicates the lack of a direct influence of one variable on the other.

\subsection{Setup: LMs as KBs}
\label{subsec:setup}

Our goal is to explain knowledge extraction from LMs, and we use the setup of `Language Models as Knowledge Bases' \cite{lama}. %
Under this setup, we sample factual knowledge triplets of (\textit{subject}, \textit{relation}, \textit{object}) from some KB (e.g. (Paris, capital-of, France)), and transform the abstract relation into a \textit{pattern} in natural language (e.g. ``[X] is the capital of [Y]''). Then, we instantiate the subject marker ([X]) with the triplet's subject, and the object marker with a masked token (e.g. ``[MASK]''), feed this \textit{cloze-pattern} to the model (e.g. ``Paris is the capital of [MASK]'') and record its predictions. If the model's prediction equals the object from the KB, we count this as a correct prediction.

\subsection{Causal Graph: Nodes and Edges}
\label{subsec:graph-explained}

We construct a causal graph describing the process discussed in Section \ref{subsec:setup}, and we present it in Figure \ref{fig:graph}.
We constructed the graph ourselves, in a process that involved multiple iterations, while reasoning about the different variables, and the causal relations between them until reaching an agreement about the final version.
The graph neatly encapsulates all of our assumptions about causal effects (and thus, also the lack of causal effects) of the variables we consider relevant.

\paragraph{Abstract Objects}
We begin with unconnected sets of \cvar{subj}, \cvar{obj} and \cvar{rel} variables, expressing subject, object and relation respectively. 
These are discrete random variables, which take the value $i$ if the $i$th subject/object/relation is sampled.
\cvar{rel} leads to different ways of expressing such relations in texts, referred to as \cvar{pattern}. Examples for the \cvar{subj}, \cvar{obj}, \cvar{rel}, and \cvar{pattern} are \cex{Paris}, \cex{France}, \cex{is-capital-of}, and \cex{[X] is the capital of [Y]}.
Together, \cvar{subj}, \cvar{obj} and \cvar{rel} generate the \cvar{KBT} (Knowledge-Base Triplet), for indicating if a specific triplet describes an event that happened in the world (true for \cex{<Paris, is-capital-of, France>}, but false for \cex{<Paris, is-capital-of, Germany>}).

\paragraph{Data Statistics}

The set of all \cvar{KBT} determines the co-occurrences between every subject-object tuple (\cvar{$SOC_{so}$}), based on their shared appearance in world's events.
It also determines the \cvar{$SO_{hC}$} variable which indicates whether an object is the highest co-occurring entity with a certain subject.

Next, combining together a \cvar{pattern}, the \cvar{KBT} triplet and the \cvar{$SOC_{so}$} --- the subject-object co-occurrence for a specific subject-object pair --- they generate textual \cvar{utterance} which then leads to a \cvar{dataset} (e.g. Wikipedia). 

The \cvar{pattern} leads to the \cvar{$POC_{uo}$} variable, which is the pattern-object co-occurrence, which determines the most cooccurring object for a specific pattern (\cvar{$PO_{hC}$}).

\paragraph{Analysis Setup}

The \cvar{dataset} is used to train a model \cvar{$\Theta$} with some objective (e.g. Masked Language Modeling for BERT \cite{bert}).
The \cvar{pattern} variable is also used to create a \cvar{cloze-pattern}, together with the \cvar{KBT}, which are used for probing a model for a specific \cvar{object}, and test if it understands that relation. In turn, that \cvar{cloze-pattern} together with the model \cvar{$\Theta$}, result in a prediction \cvar{$\hat{Y}$}.

\paragraph{Artifact Predictions}

Finally, we describe the binary variables that correspond to the hypotheses listed above, thus the \textit{artifact predictions}. %
We define three outcome variables: \cvar{$\hat{O}_{utt}$}, \cvar{$\hat{O}_{poc}$}, and \cvar{$\hat{O}_{soc}$}, each is assigned 1 if the model's prediction $\hat{Y}$ is equal to the object of the utterance \cvar{utt}, \cvar{$PO_{hC}$}, and \cvar{$SO_{hC}$}, respectively, and 0 otherwise.

\subsection{Causal Hypotheses}
After establishing the causal graph, which formulates the different variables of interest and their causal connections, we can phrase the questions from the introduction in terms of the causal effect between variables in the graph:

\begin{hypothesis}[Exact-Match]
    The \cvar{utterance} appearance affects the model's ($\Theta$) prediction $\hat{Y}$.
\end{hypothesis}

\begin{hypothesis}[Pattern-Obj Co-occurrence]
    The co-occurrence in the data between the patterns and the objects (\cvar{$PO_{hC}$}) affects the model's ($\Theta$) prediction $\hat{Y}$.
\end{hypothesis}

\begin{hypothesis}[Subject-Obj Co-occurrence]
    The co-occurrence in the data between the subjects and objects (\cvar{$SO_{hC}$}) affects the model's ($\Theta$) prediction $\hat{Y}$.
\end{hypothesis}

In what follows, we answer whether these hypotheses hold, and estimate their effect.
While a strong effect of these hypotheses may not seem problematic at first, they have serious implications for the generalization abilities of such models. 
A causal effect of the \textit{Exact-Match}\footnote{Note that the \textit{Exact-Match} hypothesis is reminiscent of the memorization definition in \citet{counterfactual_memorization}. However, while in their work they perform the ideal experiment of training multiple models on counterfactual texts, and empirically quantify memorization, we aim to estimate this effect from observational data. Since the setup is different (e.g. auto-regressive LMs vs. MLM in our case) we cannot directly compare to their results; however, such an approach can be used in future work to validate our results.} hypothesis entails that the model relies on the object that appeared with a specific utterance in the training data, and memorize it.\footnote{The \textit{exact-match} hypothesis can also be thought of as a \textit{subject-pattern-object} co-occurrence heuristic.} As such, it does not generalize the knowledge we aim to extract (the mapping between $(subj, rel) \rightarrow{} obj$), and given a paraphrase of the memorized pattern, it is likely to fail.
A causal effect of the \textit{pattern-object co-occurrence} hypothesis entails insensitive predictions of the model towards particular patterns, while disregarding the subject. As such, given the same pattern with a different subject, the model is likely to make the same prediction.
Finally, a casual effect of the \textit{subject-object co-occurrence} means the pattern, which conveys the relation, is ignored by the model. In practice, this means that for patterns expressing other relations, the model is likely to make the same prediction.

Note that these three hypotheses may be competing with one another. For instance, the \textit{pattern-object co-occurrence} and \textit{subject-object co-occurrence} are competing strategies since if a model makes use only of the subject, it means it ignores the cloze-pattern. Thus, the model relies on different hypotheses for different inputs.

\paragraph{When is reliance on heuristics a problem?} 
Features like co-occurrence can be powerful, and at times
necessary for certain reasoning and generalization
skills. Consider the statistical association that exists
between people with Italian names and the fact of being born
in Rome. While any reliance  on such an association for information that \emph{does} appear in the training data indicates that the model does not express memorized world knowledge, when it comes to querying the model on factual information that \emph{does not} appear in its training data, the model must guess. In this situation, we can expect a ``good" model to perform an educated (that is, non-random) guess, e.g., by relying on the association between the name ``Enrico Fermi" and Italy when predicting Rome as the birthplace of the physicist \cite{poerner-etal-2020-e}. 
Thus, when it comes to knowledge not included in the training data, witnessing a causal effect for statistical co-occurrence information does not rule out the fact the model expresses robust knowledge with respect to information that was included in the training data.
However, in our setup, we investigate facts from a KB based on Wikipedia, thus appearing in the training data in some textual form of the models we inspect. As such, measuring causal effect of such heuristics indicates on lack of generalization.

\section{Causal Estimation}
\label{sec:estimation}

In this section, we provide the technical background for estimating causal effects for observational data; we formalize our hypotheses based on the causal graph (\S \ref{sec:graph}) using \textit{do-calculus}, discuss the challenges and describe our solution to the question of what data to consider for estimating the causal effect.

\subsection{Estimating the Hypotheses}
We begin by formalizing our different hypotheses, and the causal estimation computation needed to calculate such effects.
Generally, the causal effect of one variable on another is described as: $P(Y|do(X))$ where $do()$ is the counterfactual function that sets the value of $X$ to a specific variable. 
In practice, we wish to compute the following equations, corresponding to the three hypotheses presented above:
\begin{align}
&P(\hat{O}_{utt}|do(\text{UTT})) \\
&P(\hat{O}_{poc}|do(\text{PO}_{\text{hC}})) \\
&P(\hat{O}_{soc}|do(\text{SO}_{\text{hC}}))
\end{align}

Indicated by the graph structure, some variables may function as confounders, and thus need to be controlled for. %
The \textit{backdoor criterion} \cite{pearl} allows us to estimate the causal effect\footnote{Note that the \textit{backdoor criterion} is not always applicable, and depends on the graph's structure.} by marginalizing on the confounder variables $Z$:

\[
P(Y|do(X)) = \sum_{z} P(Y|X, Z=z) P(Z=z)    
\]

Using the above equation, we can estimate the causal effect of one variable on another, controlling for any confounding variable from observational data alone using maximum likelihood estimates. 
Determining the variables for each equation that satisfy the \textit{backdoor criterion} can be done using the \textit{d-separation} algorithm \cite{pearl1988probabilistic}, a standard graphical models algorithm. It indicates which variables should be part of $Z$, that blocks any other information other than $X$, on $Y$ \cite{pearl}. %
We provide the formula for estimating each hypothesis by applying the aforementioned algorithm on our causal graph (Figure \ref{fig:graph}).

\paragraph{Exact-Match}

\begin{align}
    P&(\hat{y}_{utt}|do(\text{UTT})) \nonumber \\
    &= \sum_{p \in \text{PAT}} \sum_{t \in \text{KBT}} \sum_{c \in \text{SOC}_{\text{so}}} \nonumber \\
    & P(\hat{y}_{utt}|utt, \text{PAT}=p, \text{KBT}=t, \text{SOC}_{\text{so}}=\text{c}) \nonumber \\
    & \times P(\text{PAT}=p, \text{KBT}=t, \text{SOC}_{\text{so}}=\text{c})
    \label{for:em}
\end{align}

\paragraph{Pattern-Object Co-occurrence}  

\begin{align}
    P&(\hat{y}_{poc}|do(\text{\text{PO}}_{\text{hC}})) \nonumber \\
    &= \sum_{u \in utt} P(\hat{y}_{poc}|\text{PO}_{\text{hC}},\text{UTT}=u) \times P(\text{UTT}=u)
    \label{for:poc}
\end{align}

\paragraph{Subject-Object Co-occurrence}

\begin{align}
    P&(\hat{y}_{soc}|do(\text{SO}_{\text{hC}})) \nonumber \\ 
    &=\sum_{c \in \text{SOC}_{\text{so}}} P(\hat{y}_{soc}|\text{SO}_{\text{hC}},\text{SOC}_{\text{so}}=c) \nonumber \\
    & \times P(\text{SOC}_{\text{so}}=c)
    \label{for:soc}
\end{align}

\subsection{Data Population}
After formalizing the hypotheses and providing the formulas for computing each one, we are left with clear probability estimates to calculate. However, what are the data points to consider for such calculation?
In other fields, such as medicine, each patient is one
instance (also referred to as \textit{unit of analysis})
containing different features like cause and effect, as well
as confounders. However, in our scenario the scope of an individual instance is not immediately clear.

We define an individual as the (\cvar{subj}, \cvar{obj}, \cvar{rel}, \cvar{pat}) tuple.
This makes the possible population space extremely large, raising the question which individuals should be included in the population, given that large chunks of the population space are unlikely, or irrelevant. For example, consider we prompt the model with the cloze-pattern: ``Barack Obama was born in [MASK]'', it is unlikely that a good LM will predict \textit{math}. The prompted object and \textit{math} are of different types, and LMs are known to be very good at modeling entity types and selectional restrictions.
We approach the question of what population to consider by \textit{Type Preservation}, a method we developed to adjust for the object's type we consider, and \textit{Matching}, a well-established method from causality which we use to select the most similar untreated instances (i.e. where the hypothesis doesn't hold, e.g. where the subject doesn't match with the most co-occurring object in the corpus).

\paragraph{Type Preservation}
With \textit{type preservation}, we wish to avoid comparing irrelevant objects with the cloze-patterns the model is prompted with, and solely compare subject-objects of same type, which follow the relation's type. For instance, for the \textsc{born-in} relation, we only consider \textsc{Location} objects, such as \textit{Paris}, \textit{London}, etc. This is done by considering only the subject-object pairs from the same relation from the KB. 
In the subject-object co-occurrence hypothesis, since we
test whether the prediction is based on the co-occurrence,
and thus not relying on the pattern itself, we wish to
compare the predictions of additional relations where the
subject-object pair does not hold. In this case, we only consider patterns for relations that preserve the type of the object, but are not-factual. For instance, for the \textsc{born-in} relation, we may consider \textsc{died-in}, \textsc{married-in}, etc. We call these additional patterns \textit{anti-patterns}, and provide further details on their creation and statistics in Section \ref{sec:anti-patterns}.

\paragraph{Matching}
Considering solely objects that preserve the original object's type significantly reduces the number of examples, but the number of considered comparisons may still be large.
To refine the causal estimate even more, we use \textit{matching} \cite{matching-review} to balance the dataset. \textit{Matching} allows us to select control samples from the entire data pool, in such a way that for each \textit{treated} instance (where the hypothesis holds; e.g. the object co-occurs the most with some subject), we consider a \textit{control} instance, which is as similar as possible to the treated example in the confounding variables. Different methods exist for measuring similarity, and for simplicity we use the identity function for the discrete confounders, and euclidean distance for the continuous ones.

\paragraph{On the Limitations of Observational Data}
While we can use our proposed causal graph to adjust for observed confounding, unobserved confounding is the Achilles heel of most non-experimental studies, and ours is no exception. 
To perform causal inference as we do, we invoke a strong and untestable assumption, stating that all of the variables affecting both treatment and outcome are observed. Violation of this assumption, commonly known as the ignorability assumption \cite{feder2021causal} causes bias in the estimation of the effect. 

To help understand the robustness of our non-experimental findings to a potential unobserved confounder, one method involves performing a \textit{sensitivity analysis} \cite{cornfield1959smoking}. Sensitivity analysis methods deal with possible hidden confounding and attempt to measure the estimation bias under different possible models \cite{robins2000sensitivity, diaz2013sensitivity}. There are many methods for performing sensitivity analysis \cite{liu2013introduction}, which we leave to future work to explore.

\section{Experimental Design}
\label{sec:setup}

In the previous section we outlined the formulas that allow us to estimate the causal effect from observational data, and the filters that define the population of interest.
In this section we detail our experimental setup for calculating the causal effects. 
Overall, the objective is to convert the different data sources into tables, one for each hypothesis, which will be used to calculate the probabilities of interest.
We do so by combining the different data sources, such as the (subject, relation, object) triplets, the patterns that correspond to each relation, the prediction for the cloze-patterns, the data statistics (such as the subject-object co-occurrence), etc. Once we obtain a corresponding table for each hypothesis we can estimate the probabilities of each formula based on the causal estimation described in the previous section. 

\subsection{From Theory to Practice}
\label{subsec:theory-practice}

We begin by describing the corresponding populations for each hypothesis. 
The individual we consider in each population is composed of the (\cvar{subj}, \cvar{obj}, \cvar{rel}, \cvar{pat}) tuple.
Then, for each hypothesis we add the relevant features for computing the corresponding hypothesis (e.g. the co-occurrence counts between the subject and the object).

It can be useful to think about this process as building a table, where each row corresponds to an instance, with the (\cvar{subj}, \cvar{obj}, \cvar{rel}, \cvar{pat}) tuple as the defining instance, and other columns as additional features, which allow us to estimate the causal effects.

\paragraph{Exact-Match}

\begin{table*}[t!]
    \centering
\resizebox{1\textwidth}{!}{%
\begin{tabular}{llllrllll}
\toprule
             Subj &      Obj &                Rel &                                  Pattern &     $SOC_{so}$ &    $SOC_{so} (bin)$ &   Utterance &   $\hat{Y}$ &     $\hat{O}_{utt}$ \\
\midrule
      True Detective  &        HBO  &    originally-aired-on  &                       [Y] released [X].  &             116  &           L  &         True  &      Netflix  &             False  \\
       True Detective  &        HBO  &    originally-aired-on  &                    [Y] is to debut [X].  &             116  &           L  &        False  &      Netflix  &             False  \\
  The Big Bang Theory  &        CBS  &    originally-aired-on  &        [X] was originally aired on [Y].  &             200  &           L  &         True  &          NCB  &             False  \\
  The Big Bang Theory  &        CBS  &    originally-aired-on  &                        [Y] debuted [X].  &             200  &           L  &        False  &          NCB  &             False  \\
  \midrule
             Edmonton  &    Alberta  &             is-capital  &         [X] is the capital city of [Y].  &            7147  &          XL  &         True  &      Alberta  &              True  \\
             Edmonton  &    Alberta  &             is-capital  &    [Y], which has the capital city [X].  &            7147  &          XL  &        False  &      Alberta  &              True  \\
             Jayapura  &      Papua  &             is-capital  &         [X] is the capital city of [Y].  &             112  &           L  &         True  &    Indonesia  &             False  \\
             Jayapura  &      Papua  &             is-capital  &         The capital city of [Y] is [X].  &             112  &           L  &        False  &        Nepal  &             False  \\
\bottomrule
\end{tabular}

}
\caption{A subset of the table for computing the causal effect for the \textit{exact-match} hypothesis.}
\label{tab:memorized-example}
    
\end{table*}

\begin{table*}[t!]
    \centering
\resizebox{1\textwidth}{!}{%

\begin{tabular}{llllllll}
\toprule
      Subj &  Obj &                     Rel &  Pattern &  $UO_{hC}$ &           Utterance & $\hat{Y}$ &  $\hat{O}_{uoc}$ \\
\midrule
        Daria  &       MTV  &    originally-air-on  &            [X] debuted on [Y].  &               True  &        False  &              MTV  &        True  \\
        Daria  &       BBC  &    originally-air-on  &            [X] debuted on [Y].  &               False  &        False  &              MTV  &        False  \\

        The NFL Today  &      CBS   &    originally-air-on  &               [Y] debuted [X].  &               True  &         True  &             ESPN  &       False  \\
        The NFL Today  &      NBC   &    originally-air-on  &               [Y] debuted [X].  &               False  &         False  &             ESPN  &       False  \\
               
                 \midrule
                 
                       Paris  &    France  &           is-capital  &     [X] is the capital of [Y].  &               True  &         True  &           France  &        True  \\
                       Paris  &      West  &           is-capital  &     [X] is the capital of [Y].  &              False  &        False  &           France  &       False  \\
                      Ankara  &    Serbia  &           is-capital  &       [X], the capital of [Y].  &               True  &        False  &           Turkey  &       False  \\
                      Ankara  &    Uganda  &           is-capital  &     [X], the capital of [Y].  &                 False  &        False  &           Turkey  &       False  \\
\bottomrule
\end{tabular}

}
\caption{A subset of the table for computing the causal effect for the \textit{pattern-object co-occurrences} hypothesis. The $SOC_{so} (bin)$ variable is computed based on the bins defined in Section \ref{subsec:theory-practice}.}
\label{tab:default_object-example}
    
\end{table*}

\begin{table*}[t!]
    \centering
\resizebox{1.\textwidth}{!}{%
\begin{tabular}{llllrllll}
\toprule
Subj & Obj & Rel & Pattern & $SOC_{so}$ & $SOC_{so} (bin)$ & $SO_{hC}$ & $\hat{Y}$ & $\hat{O}_{soc}$ \\
\midrule

 Safari  &     Apple  &   developed-by  &        [X] is a product of [Y].  &         269  &        L  &         True  &     Apple  &     True  \\
 Safari  &    Google  &   developed-by  &        [X] is a product of [Y].  &         256  &        L  &        False  &     Apple  &    False  \\
 Safari  &     Apple  &   developed-by  &            [X] was sold to [Y].  &         269  &        L  &         True  &    Boeing  &    False  \\
 Safari  &    Google  &   developed-by  &            [X] was sold to [Y].  &         256  &        L  &        False  &    Boeing  &    False  \\
 
 \midrule
 
  Paris  &    France  &     capital-of  &         [X], the capital of [Y]  &       31535  &       XL  &         True  &    France  &    False  \\
  Paris  &   Germany  &     capital-of  &         [X], the capital of [Y]  &        3042  &       XL  &        False  &    France  &    False  \\
  Paris  &    France  &     capital-of  &   [X] is not the capital of [Y]  &       31535  &       XL  &         True  &    France  &     True  \\
  Paris  &   Germany  &     capital-of  &   [X] is not the capital of [Y]  &        3042  &       XL  &        False  &    France  &     False \\

\bottomrule
\end{tabular}

}
\caption{A subset of the table for computing the causal effect for the \textit{subject-object co-occurrences} hypothesis. The $SOC_{so} (bin)$ variable is computed based on the bins defined in Section \ref{subsec:theory-practice}.
}
\label{tab:cooccurrence-example}
    
\vspace{-0.1in}
\end{table*}

In this setup, we use all triplets from the KB (meaning that these triplets are factually correct), and combine them with all of the paraphrases for each relation (obtained by the cartesian product between the KB triplets and the paraphrase list, per relation).
For each instance we add the information about the $SOC_{so}$ - the subject-object co-occurrence in the training data, their binned version, and whether the $utt$ (the instantiation of the pattern with the subject and object) appeared in the data or not. Finally, we add the model's prediction on the cloze-pattern, and whether it matches the hypothesis (\cvar{$\hat{O}_{utt}$}).
We keep all instances that appeared in the training data, and we match these, based on the confounders, meaning that we use the same KBT, but another pattern. An example of the described population is presented in Table \ref{tab:memorized-example}.

\paragraph{Pattern-Object Co-occurrence}  

In this setup, we consider all (subject, object) pairs for each relation. This means that we include both pairs that hold for a relation (e.g. Paris, is-capital, France), and such that do not hold (e.g. Ankara, is-capital, Serbia).
For each instance we add information about whether the object is the most cooccurring with the pattern ($PO_{hC}$), whether the utterance appeared in the training data ($utt$), the model's prediction and whether it matches the hypothesis ($\hat{O}_{poc}$).
Out of the entire population, we keep instances whose pattern-object corresponds to the most-common object for that pattern, and match those with an instance with the same subject-pattern, where the object is the next most frequent with the pattern. We only keep instances where the frequency of the pattern-object is higher than 5. An example of the described population is presented in Table
\ref{tab:default_object-example}.

\paragraph{Subject-Object Co-occurrence}

In this last setup, we also consider all (subject, object) pairs per relation, but not only the paraphrased patterns for each relation with all (subject, object) pairs, but also the \textit{anti-patterns}, which preserve the subject-object type, but represent a different relation (\S \ref{sec:anti-patterns}).
For each instance we add the information about the co-occurrence between the subject and object ($SOC_{so}$), whether the object is the most frequent for that subject ($SO_{hC}$), the model's prediction ($\hat{Y}$) and whether it matches the hypothesis ($\hat{O}_{soc}$).
Since the magnitude of the co-occurrence is most likely to influence the prediction rather than the exact number, and in order to cope with the sparsity of this variable, we group the values into 5 bins in the following ranges: $[0,1]$, $(1,10]$, $(10,100]$, $(100,1000]$, and $(1000, \infty]$, named XS, S, M, L, and XL, respectively.
In the final population, we keep all instances where the object co-occurs the most with a subject, and match these by selecting the next most common object, with the same subject and pattern.
An example of the described population is presented in Table
\ref{tab:cooccurrence-example}.

\subsection{Data collection}

We now describe the different variables from the graph that are required for estimating the effects of the different hypotheses.
\paragraph{Abstract Objects}
We consider the subject, object and relation triplets from T-REx \cite{elsahar2018t}, part of LAMA \cite{lama}, that was additionally filtered by \citet{pararel}.
The KBT value is defined by whether such triplet appears in the KB or not. The pattern variables take the different textual patterns from \textsc{ParaRel} \cite{pararel}.
In addition, we build an anti-pattern dataset where for each relation, we construct patterns that maintain the type of the corresponding objects (e.g. \textit{location}), but the answer is likely to be different. For instance, for the pattern ``X worked for Y'', we create a pattern ``X acquired Y''. We provide more details on this data collection in Section \ref{sec:anti-patterns}.

\paragraph{Analysis Setup}

The \textit{cloze-pattern} are instantiations of the \textit{patterns}, where a subject position is replaced with a subject from the KB, and the object is replaced with a masked token.
The training data is based on what the models were trained on, thus in our case it's English Wikipedia and the Book Corpus \cite{book}.

Finally, the prediction variable is calculated by feeding the cloze-patterns to the model. Following \citet{Xiong2020Pretrained,ravichander-etal-2020-systematicity,kassner2021multilingual,pararel}, we restrict the candidate sets to the set of gold objects from the same relation, in order to avoid potentially correct, but non-factual completions of the LM (e.g. a prediction of \textit{TV} for the pattern ``True Detective was originally aired on [MASK]'').

\paragraph{Data Statistics}

To collect statistics on the variables related to the training data, we use SPIKE \cite{spike}, a syntactic search engine that allows fast syntactic search across an indexed corpus of text.
We jointly index Wikipedia and the Book Corpus to query the relevant information.
We assign corresponding values to the following variables:

\indent \cvar{$utt$}: For each \cvar{$utterance$} from the set of considered instantiated patterns, we use the index and check whether it appeared in the corpus or not.\\
\indent \cvar{$SOC_{so}$}: For each subject-object pair from the corpus, we count the number of times they appear in the same sentence.\\
\indent \cvar{$SO_{hC}$}: This variable is deterministically derived from  \cvar{$SOC_{so}$}, and is assigned ``true'' iff the object appears the most with a subject.\\
\indent \cvar{$POC_{uo}$}: This variable is deterministically derived from all possible \cvar{pat} that appear in the data, by removing the subjects, and counting the number of times each repeats.\\
\indent \cvar{$PO_{hC}$}: This variable is deterministically
derived from  \cvar{$POC_{uo}$}, and is assigned ``true'' iff a particular object appears the most with a pattern.

\paragraph{Artifact Predictions}

Finally, the artifact prediction 
variables are simply calculated by comparing the model's prediction, with the relevant hypothesis. In the exact-match hypothesis, we compare the prediction to the object that appeared with the considered utterance. In the pattern-object co-occurrence we compare the prediction with the co-occurring object, and in the subject-object co-occurrence, we compare the prediction with the considered object.

\begin{table*}[t!]
    \centering
\resizebox{1\textwidth}{!}{%
\begin{tabular}{llll}
\toprule
Relation & Pattern & Anti-Pattern \#1 & Anti-Pattern \#2 \\
\midrule

country & [X] is located in [Y]. & [X] is located next to the border with [Y]. & [X] was constructed outside of [Y]. \\
continent & [X] is located in [Y]. & [X] is not located in [Y]. & [X] is mistakenly believed to be in [Y]. \\
field of work & [X] works in the field of [Y]. & [X] invented the field of [Y]. & [X] has never gotten to work in the field of [Y]. \\
genre & [X] plays [Y] music. & [X] invented [Y] music. & [X] never listens to [Y] music. \\
location & [X] is located in [Y]. & [X] is located in [Y]'s twin city. & [X] has recently moved to [Y] from Tel Aviv. \\
original network & [X] was originally aired on [Y]. & [X] was bought by [Y]. & Writers from [Y] wrote the series [X]. \\

\bottomrule
\end{tabular}

}
\caption{Examples of patterns for different relations, and the corresponding \textit{anti-patterns} we collected in this work. Each \textit{anti-pattern} is some modification of the original pattern, which changes the meaning of the original pattern, such that the answer's type would remain the same, but the answer is likely to change.}
\label{tab:anti-patterns}
    
    \vspace{-0.1in}
\end{table*}

\subsection{Collecting Anti-Patterns}
\label{sec:anti-patterns}

We wish to test the hypothesis that a model ignores the given pattern, and mainly makes use of the subject to predict an object. We thus need to provide some patterns which are not paraphrases of one another.
One option to do so is to take patterns from other relations. However, this solution has two issues: First, some subject-object pairs may hold for different relations (e.g. there's a non-negligible probability that a person was born and worked in the same country). Second, we found that modern language models have a good type-inference capability - that is, when given a prompt that entails a specific class of answers (e.g. a location for the born-in relation), they tend to provide location predictions. Thus, swapping subject/object pairs with non-matching type relations, is problematic.

For these reasons, we construct  an \textit{anti-pattern}
resource, where for each relation, we write patterns
expressing different relations, that matches the type
(e.g. location), but are also unlikely to have the same
answer. We refer to these patterns as \textit{anti-patterns}.
The resource contain 194 \textit{anti-patterns} for 35 relations, which were constructed by one of the authors, and verified by another. Some examples are presented in Table \ref{tab:anti-patterns}.
Note that for the \textit{anti-patterns}, the KBT value is set to false.

\subsection{Models}
We experiment with 32 different models, spanning across three models families from the Masked Language Models (MLM) family,\footnote{It would be interesting to experiment with other PLM architectures such as T5 \cite{t5}, however, they require a different evaluation setup, which we leave for future work.} namely: BERT, RoBERTa and ALBERT. Notice that our analysis validity depends on having access to the training data of the inspected models, as statistics such as the co-occurrences between entities are directly computed from the data. Therefore, our analysis can't be done on models who's training data was not released, emphasizing the importance of releasing such information to the community.

We experiment with both BERT variants: base and large \cite{bert}, the MultiBERTs \cite{sellam2021multiberts}, a collection of 25 BERT-base models, trained using similar hyperparameters and on the same data, but using different random initialization and data shuffling.
This allows us to provide more rigorous results, and provide evidence that the inspected hypotheses are independent on random seeds.

In addition, we also experiment with the four size variants of ALBERT \cite{albert}, a BERT-like model, with smaller embedding size, and parameter sharing, which makes it much more efficient in parameters, and outperforms BERT on a range of tasks.\footnote{We use ALBERT v1, that was trained on the same data as BERT.}

Finally, we also experiment with the base version of RoBERTa \cite{roberta}. Since the original training data of RoBERTa is not publicly available, we cannot analyze the released model. Instead, we retrained a version of this model using the same training data of BERT (Wikipedia and the book corpus, which we have access to).
We trained the model for 99,650 steps, corresponding to 83 epochs over the data, reaching 3.62 perplexity over a subset of Wikipedia. Unless mentioned otherwise, we report the results of this model using the last checkpoint, after 83 epochs.
To verify our model was properly trained, we fine-tune it over SQuAD1.1 \cite{squad} for 10 epochs, and reach 89.6 F1 score on the development set, which is comparable to the 88.5 F1 that BERT-base achieved while trained on the same data.
We save the checkpoints of this model after each epoch, to perform additional analysis on the training dynamic of such a model, and release the checkpoints to the community at \url{https://huggingface.co/yanaiela}.\footnote{All experiments involving PLMs were run through the HuggingFace library \cite{hf}.}

\subsection{Metric}

We report our results using the average treatment effect \cite[ATE,][]{pearl}, which is the mean difference between instances from the treatment to the control group (in our case, the subject-object pairs that co-occurred the most in the training data, vs. the pairs that didn't, for instance).  %
Formally, it can be expressed as follows: $E(Y|do(X=0))-E(Y|do(X=1))$, where $Y$ and $X$ are the outcome, and cause variables, respectively. The ATE values range between -1, and 1, where positive and negative values mean the effect is positive, and negative respectively. Small values around zero mean that the effect is negligible, thus, allowing researchers to delete the drawn edge, which signifies that the variables are not causally related.

To compute ATE, we estimate the causal effect of each hypothesis twice: once when the treatment is `used', and once when it's not (as we operate under a binary treatment scenario, this simply means that the values are set to $1$ or $0$). We calculate each condition using Equations \ref{for:em}-\ref{for:soc}, and subtract the results.
Figure \ref{fig:intro_fig} provides an intuition about this measure; we are after the result of the simulated world, where an intervention of a counterfactual occurred. In the figure, we ask about the hypothetical situation where the object \textit{Chicago} didn't appear the most with \textit{Barack Obama}, and the result of the same exact model that was trained on such hypothetical dataset.

\section{Results}
\label{sec:restuls}
\begin{table*}[t!]
    \centering
\begin{NiceTabular}{llll}
\toprule
       Model & Exact Match & POC & SOC \\
\midrule

 BERT-base &          2.95 &           12.42 &            18.54 \\
  BERT-large &        4.14 &          9.27  &            19.81 \\
  MultiBERTs &       3.73$\pm$1.18 &  9.12$\pm$1.16 &    17.83$\pm$0.58 \\
  
  \midrule
  
  ALBERT-base &           5.38 &             3.1 &           14.51 \\
   ALBERT-large &           6.32 &            8.15 &           13.95 \\
  ALBERT-xlarge &           4.24 &            6.69 &           15.18 \\
 ALBERT-xxlarge &           4.89 &            6.47 &           16.52 \\
 
 \midrule
  
  RoBERTa-base* &         5.39        & 16.62 & 5.83 \\
 
\bottomrule
\end{NiceTabular}

\caption{ATE results of the three hypotheses (\textit{Exact-Match}, \textit{Pattern-Object Co-occurrence}, and \textit{Subject-Object Co-occurrence}) for the different models we consider. For the MultiBERTs, we report the mean and std over the 25 models. 
}
\label{tab:ate-results}
    
\vspace{-0.1in}
\end{table*}

After calculating the relevant tables for each hypothesis, we estimate the causal effect using Equations \ref{for:em}-\ref{for:soc}.
The results are displayed in Table \ref{tab:ate-results}.

\subsection{Main Results}

\paragraph{BERT}
Overall, we found that all three hypotheses have an effect on BERT's predictions. The ATE for BERT-base on exact-match, pattern-object co-occurrence and the subject-object co-occurrence are 2.95, 12.42, and 18.54 respectively, and for BERT-large 4.14, 9.27, and 19.81 respectively. 
This means, for instance, that 18.54\% of BERT-base's predictions will be based on the most co-occurring object with some subject.
Interestingly, the co-occurrence effect is the strongest one out of the tested hypotheses. This shows that such a simple statistic greatly affects the model predictions. Recall that since we included both factual prompts, and non-factual prompts (the anti-patterns) the reliance of such co-occurrences is high, which questions the claims of factual knowledge encoded by BERT \cite{lama,petroni2020context}.

Noticeably, the memorization effects of the exact-match experiments are relatively low, compared to the other effects. While the setup is not directly comparable to previous works, similar trends have been observed in the literature \cite{carlini2020extracting,carlini2022quantifying_memorization,counterfactual_memorization}.
One aspect to consider is the strict exact-match definition.
Considering a more relaxed definition (e.g. one that allows minor matching strategies, as allowing differences in punctuation), may reveal stronger effects of this strategy. 
However, due to the nature of texts which allows a great deal of diversity in expressing the same information, we do not consider small differences such as the inclusion of a comma as the same utterance. A more relaxed definition should be considered in future work, which may find stronger effects. 

\paragraph{MultiBERTs}

Next, we report the average and std of the 25 MultiBERTs.
The average ATE across the MultiBERTs are 9.12, 3.73 and 17.83 for the three hypotheses, respectively, with small standard deviations (0.58-1.18). These results are similar to the other BERT models, strengthening the findings that such heuristics are indeed being used by this architecture.

\paragraph{ALBERT}
We report the results over the four different ALBERT sizes that were released.
The causal effects of the heuristics on these models are still strong, however, there are some differences from the BERT models.
First, in the base model, the ATE of the pattern-object co-occurrence is lower than that of the exact-match heuristic. Second, there's no clear trend of the model size on the heuristics' effects, contrary to the increased sensitivity to gender bias by larger models, observed by \citet{vig2020causal}.

\paragraph{RoBERTa}

Interestingly, our re-implementation of RoBERTa behaves differently from the other models. Specifically, the influence of the pattern-object co-occurrence and the subject-object co-occurrence switched, achieving 16.62 and 5.83, respectively. The main difference between RoBERTa's architecture vs.\  BERT and ALBERT is the lack of an inter-sentence task (next sentence prediction for BERT, sentence order prediction for ALBERT), which may be the source of such a difference between the models.

\subsection{Training Dynamics of the Heuristics}
\begin{figure}[t]
\centering
\includegraphics[width=1.\columnwidth]{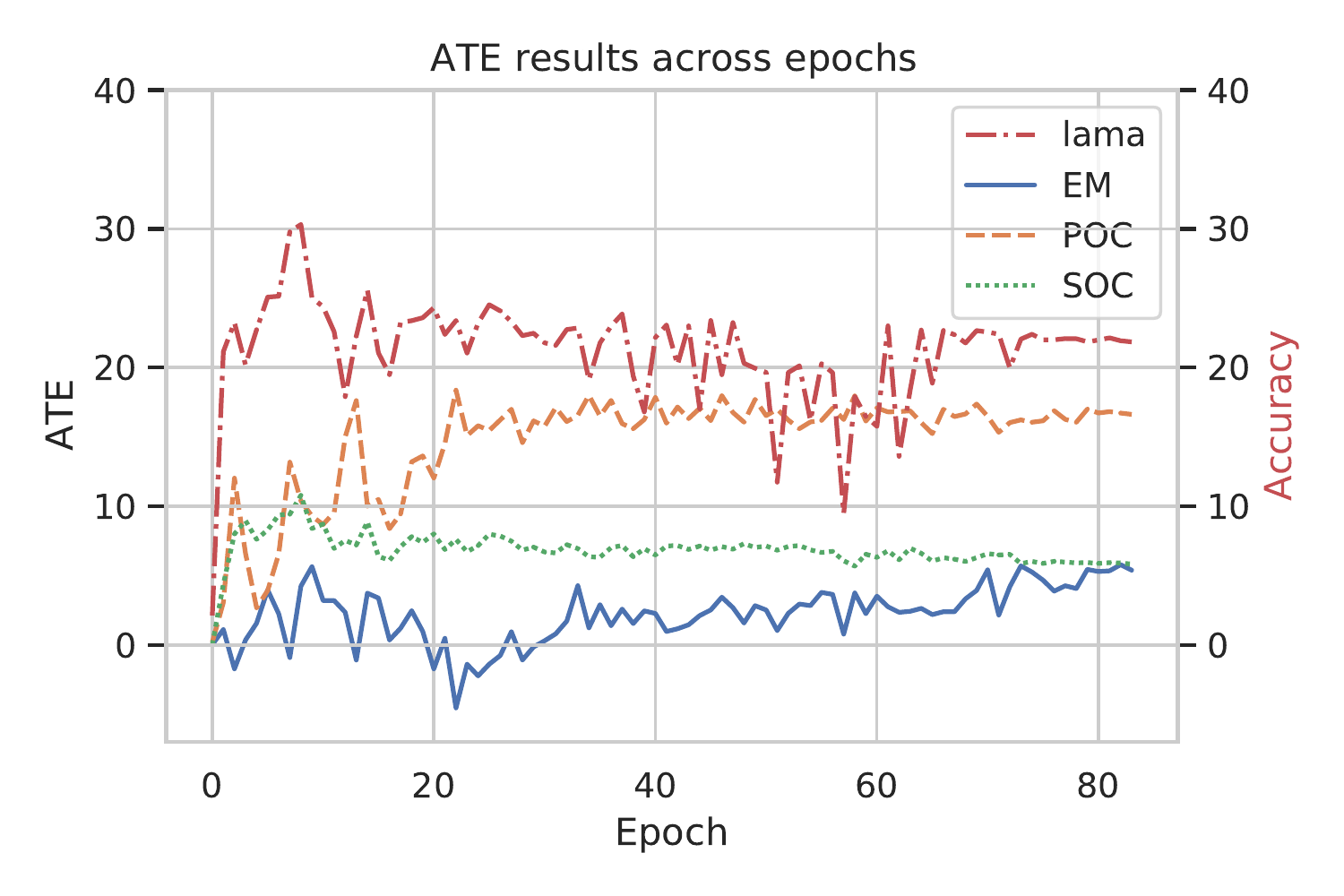}
\caption{
ATE results across epochs of our RoBERTa-base model. We report the results of the three heuristics we consider, as well as the LAMA accuracy results over the T-REx relations we consider over 84 epochs.}
\label{fig:over-time}
\vspace{-0.2in}
\end{figure}
In all experiments so far we focused on a single model checkpoint -- the final one. However, what happens during the training of such LMs in terms of adoption or abandonment of the heuristics of interest? To answer this question, we use the RoBERTa-base model we trained, and apply our framework after each checkpoint (84 in total, including the randomly initialized model). Our results are plotted in Figure \ref{fig:over-time}.
We also plot the results of the LAMA probe \cite{lama}, on the relations we use from T-REx. 

While the LAMA results oscillates around 22\% accuracy (with some large outliers, above 30\% and below 10\% accuracy), the different heuristics behave differently. The exact-match scores are rather low, and oscillates around 0\%, but they slowly increase after 30 epochs and reach 5.39\% by the last epoch. This behavior can intuitively be explained by the continual overfitting of the model, that is also seen by the improved perplexity over the training data.
Finally, the utterance and subject-object co-occurrence heuristics stabilize after 20 epochs.

\begin{table*}[t!]
    \centering
\begin{NiceTabular}{cllll}
\toprule
      & Model & Exact Match & POC & SOC \\
      
\midrule

\parbox[t]{2mm}{\multirow{2}{*}{\rotatebox[origin=c]{90}{Bl}}} &
    Heuristic   &      100 & 100 & 100 \\
  & Perfect     &      1.62 & 0 & 0.23 \\
  
  \midrule
 
\parbox[t]{2mm}{\multirow{8}{*}{\rotatebox[origin=c]{90}{Random Weights}}} &
  BERT-base &      -0.45        & -2.88 &             0.29 \\
 & BERT-large &    0.02         & -3.52 &             -0.12 \\
 & MultiBERTs &    0.0$\pm$0.78 & 0.44$\pm$2.22 &     0.14$\pm$0.25 \\
 
 \cmidrule{2-5}
 
 & ALBERT-base &          -0.77 &            0.25 &            0.47 \\
 & ALBERT-large &           0.08 &           -2.58 &           -0.07 \\
 & ALBERT-xlarge &          -0.03 &            1.08 &            0.06 \\
 & ALBERT-xxlarge &          -0.04 &            3.58 &            0.48 \\

 \cmidrule{2-5}

 & RoBERTa-base & 0.03 & 0.20 & 0.00 \\

\bottomrule
\end{NiceTabular}

\caption{Control ATE results of the three hypotheses (\textit{Exact-Match}, \textit{Pattern-Object Co-occurrence}, and \textit{Subject-Object Co-occurrence}). 
We report the results for the two baselines at the top, and the randomly initialized models results at the bottom.
The first two (\textit{Bl} in the upper part), consist of a model that always uses the heuristic, and a model that always predicts the correct object (based on the KB). The random weights models (\textit{Random Weights} in the lower part), are randomly initialized, and thus, neutralizing the effect of the data on the model.
}
\label{tab:control-results}
    
\end{table*}

\subsection{Causal Effect per Relation}

\begin{table}[t!]
    \centering
\resizebox{0.8\columnwidth}{!}{%
\begin{tabular}{clr}
\toprule
                            & Pattern &    CATE \\
\midrule
 
 \parbox[t]{2mm}{\multirow{6}{*}{\rotatebox[origin=c]{90}{Subj-Obj Cooc}}} 
                           & occupation &   0.01 \\
                       & place of birth &   1.93 \\
                        & field of work &   1.96 \\
                   \cmidrule{2-3}
                            & continent &  50.78 \\
                           & capital of &  57.64 \\
                         & manufacturer &  57.73 \\
                        \midrule
                        \midrule
\parbox[t]{2mm}{\multirow{6}{*}{\rotatebox[origin=c]{90}{Pat-Obj Cooc}}}  
                              & part of &   0.00 \\
               & country of citizenship &   0.02 \\
                        & field of work &   0.28 \\
                    
                    \cmidrule{2-3}
            & original language of show &  27.01 \\
                                & genre &  50.31 \\
                            & member of &  50.34 \\
                            \midrule
                            \midrule
\parbox[t]{2mm}{\multirow{6}{*}{\rotatebox[origin=c]{90}{Exact Match}}}  
            & original language of show &   0.00 \\
                            & developer &  -0.22 \\
                            & continent &  -0.57 \\
                    \cmidrule{2-3}
                            & member of &  20.46 \\
                                & genre &  30.89 \\
                             & religion &  42.00 \\
\bottomrule
\end{tabular}

}
\caption{Top-3 most and least strong mean effects (measured by CATE) of the MultiBERTs models (averaged over the 25 models), out of all the patterns, for each hypothesis.}
\label{tab:cate-results}
    
\vspace{-0.1in}
\end{table}

Next, we experiment with the causal effect measured per relation individually;
instead of measuring the effect on the entire population at hand, we condition on one population at a time and measure the ATE for each relation individually (also known as the Conditional Average Treatment Effect (CATE)).
We report the results of these experiments in Table \ref{tab:cate-results}. For each hypothesis, we ran the experiment on all relations, and showcase the 3 most, and least strong effects, measured on the mean of the 25 MultiBERTs models.

Interestingly, the strongest effects for each hypothesis are high: 42.0, 50.34 and 57.73 for the exact-match, pattern-object co-occurrence, and subject-object co-occurrence respectively. The lowest are all close to zero.
Another interesting observation is the trade-off of such heuristics. For instance, the \textsc{capital-of} relation who's CATE is high for the subject-object co-occurrence hypothesis (57.64), is almost non-existing for the pattern-object hypothesis (0.41). 
Notice, that while for for some of the hypotheses the ATE (over the entire relations) are low, by using CATE we observe strong effects per relation, e.g. 42.0 for the religion relation with the exact-match hypothesis.

We consider multiple reasons for the different effects between relations. First, while the model had access to all of the relation instances we inspect, there are no guarantees that the model retained them all. As such, the CATE of an hypothesis may be small, due to the lack of factual knowledge. Moreover, while we provide three hypotheses explaining the prediction, there are more possible explanations for the model's predictions which we do not consider. Finally, it is also possible that the model acquired certain facts that it generalizes to, which would make the effect small.

\subsection{Controls}

We provide three baselines to better interpret our findings.
First, we conduct two simple baselines: Heuristic and Perfect. The first, Heuristic, simply uses the corresponding heuristic for prediction. For instance, in subject-object co-occurrence (SOC), it always predicts the object that co-occurs the most with the given subject. As such, by definition, the ATE score for such an experiment is always 100. The second, Perfect, is an oracle baseline, which always predicts the correct answer (from the KB). This baseline is interesting since at times, the correct answer may be aligned with some heuristic. However, due to our experimental design, the results are low (1.62 for Exact-Match, 0 for pattern-object co-occurrence, and 0.23 for subject-object co-occurrence).

In addition to the baselines above, we also experimented with models whose weights are randomly initialized, to provide another controlled experiment. In this case, where the models were not trained on any data, we expect the effect to be much smaller than the trained models. And indeed, as can be seen in the second part of Table \ref{tab:control-results}, most results are close to zero (except for the pattern-object co-occurrence where the ATE is negative, with -2.88 and -3.52 for BERT-base and large, and for ALBERT-large and xxlarge, with -2.58, and 3.58 respectively). %
These results strengthen our findings, showcasing the effects are likely to be caused indeed by the hypotheses we consider.

\section{Discussion and Conclusions}
\label{sec:conclusions}

In this work, we investigate the influence of co-occurrence statistics from the training data on model predictions. We propose a methodology for establishing a framework that estimates causal effects form observational data applied in NLP.
Then, we use this framework for discovering the effect of superficial co-occurrence statistics on model predictions, namely, \textit{Exact-Match}, \textit{Pattern-Object Co-occurrence}, and \textit{Subject-Object Co-occurrence} on 32 different models. We find that such heuristics causally affect BERT-like model predictions, indicating sub-optimal generalization of these models.

We believe that causal-based approaches are crucial for understanding and interpreting neural networks, and language models in particular, by providing the language and tools to answer such questions.
While in this work we tackled a particular analysis setup, there are many challenges and questions left open. For instance, we found that the inspected heuristics are being used to some degree, but what other heuristics do these models employ, and how much of such knowledge can be extracted robustly?
Moreover, while we constructed the causal graph based on our expertise in the field, its exact structure may be left for refinement and discussion for future work.

\section*{Acknowledgements}
We would like to thank the Katherine A. Keith, Zhijing Jin and Sian Gooding for helpful discussions and comments on this paper. In addition, we thank Stella Bidermann for providing us the computational resources for training RoBERTa-base.

\bibliography{tacl2021}
\bibliographystyle{acl_natbib}

\end{document}